\documentclass[twocolumn, switch]{article} 

\usepackage{preprint}

\usepackage{amsmath, amsthm, amssymb, amsfonts}

\usepackage[numbers,square]{natbib}
\usepackage[utf8]{inputenc}	
\usepackage[T1]{fontenc}	
\usepackage{xcolor}		
\usepackage[colorlinks = true,
            linkcolor = purple,
            urlcolor  = blue,
            citecolor = cyan,
            anchorcolor = black]{hyperref}	
\usepackage{booktabs} 		
\usepackage{nicefrac}		
\usepackage{microtype}		
\usepackage{lineno}		
\usepackage{float}			

\usepackage{lipsum}		

\usepackage{newfloat}
\DeclareFloatingEnvironment[name={Supplementary Figure}]{suppfigure}
\usepackage{sidecap}
\sidecaptionvpos{figure}{c}

\usepackage{titlesec}
\titlespacing\section{0pt}{12pt plus 3pt minus 3pt}{1pt plus 1pt minus 1pt}
\titlespacing\subsection{0pt}{10pt plus 3pt minus 3pt}{1pt plus 1pt minus 1pt}
\titlespacing\subsubsection{0pt}{8pt plus 3pt minus 3pt}{1pt plus 1pt minus 1pt}

\usepackage{tikz,xcolor,hyperref}

\definecolor{lime}{HTML}{A6CE39}
\DeclareRobustCommand{\orcidicon}{
	\begin{tikzpicture}
	\draw[lime, fill=lime] (0,0) 
	circle [radius=0.16] 
	node[white] {{\fontfamily{qag}\selectfont \tiny ID}};
	\draw[white, fill=white] (-0.0625,0.095) 
	circle [radius=0.007];
	\end{tikzpicture}
	\hspace{-2mm}
}
\foreach \x in {A, ..., Z}{\expandafter\xdef\csname orcid\x\endcsname{\noexpand\href{https://orcid.org/\csname orcidauthor\x\endcsname}
			{\noexpand\orcidicon}}
}

\title{Scaffold Embeddings: Learning the Structure Spanned by Chemical Fragments, Scaffolds and Compounds}

\usepackage{xwatermark}
\newwatermark[firstpage,color=gray!90,angle=0,scale=0.28, xpos=0in,ypos=-5in]{*correspondence: \texttt{aclyde@uchicago.edu}}

\usepackage{authblk}

\author[1,2\thanks{\tt{aclyde@uchicago.edu}}]{Austin Clyde\orcidA{}}
\author[2,3]{Arvind Ramanathan\orcidB{}}
\author[1,4]{Rick Stevens}

\affil[1]{Department of Computer Science, University of Chicago, Chicago, IL 60637}
\affil[2]{Data Science and Learning Division, Argonne National Laboratory, Lemont, IL 60439}
\affil[3]{University of Chicago Consortium for Advanced Science and Technology, Chicago, IL 60637}
\affil[4]{Computing, Environment, and Life Science Directorate, Argonne National Laboratory, Lemont, IL 60439}

\newcommand{\Expand}{\texttt{Expand}$_\Phi$}
\newcommand{\Successor}{\texttt{Successor}$_\Phi$}
\newcommand{\Intersection}{\texttt{Intersection}}
\newcommand{\Union}{\texttt{Union}$_\Phi$}
\newcommand{\Predecessor}{\texttt{Predecessor}}
\newcommand{\Scaffold}{\texttt{Scaffold}}
\newcommand{\Mspace}{\mathcal{M}}
\newcommand{\Sspace}{\mathcal{S}}

\usepackage{subfiles}
\begin{document}

\twocolumn[ 
  \begin{@twocolumnfalse} 
  
\maketitle

\begin{abstract}
Molecules have seemed like a natural fit to deep learning’s tendency to handle a complex structure through representation learning, given enough data. However, this often continuous representation is not natural for understanding chemical space as a domain and is particular to samples and their differences. We focus on exploring a natural structure for representing chemical space as a structured domain: embedding drug-like chemical space into an enumerable hypergraph based on scaffold/fragment classes linked through an inclusion operator. This paper shows how molecules form classes of scaffolds, how scaffolds relate to each in a hypergraph, and how this structure of scaffolds is natural for drug discovery workflows such as predicting properties and optimizing molecular structures.  We compare the assumptions and utility of various embeddings of molecules, such as their respective induced distance metrics, their extendability to represent chemical space as a structured domain, and the consequences of utilizing the structure for learning tasks.
\end{abstract}
\vspace{0.35cm}

  \end{@twocolumnfalse} 
] 



\section{Introduction}
The enormous design space of chemical compounds, estimated to be about $10^{60}$~\cite{bohacek1996art}, motivates an immediate need for efficient and often automated exploration for synthesis and assay development for various applications, including drug discovery and materials design. Computational enumeration of chemical space is a long-studied problem since the early ages of computing \cite{cernak2018machine}. The current state of the art projects have enumerated around 2 billion drug-like compounds, and GDB has around 166 billion compounds of up to 17 atoms of C, N, O, S, and halogens \cite{patel2020savi, ruddigkeit2012enumeration}. Even with these vast libraries, recent work has shown a vast difference between the diversity enumerated in ultra-large libraries and the underlying space \cite{jia2019anthropogenic}. Especially in the context of drug discovery, an emerging need in the cheminformatics community is the ability to  \emph{navigate} this enormous design space in the hopes of generating new molecules (or designs) that can optimally bind to a protein/drug-target of interest or \emph{refine} molecules based on specific physio-chemical and safety features that make it attractive as a drug that can be formulated for the market. 

Given the vastness of drug-like chemical space, how can we computationally explore it? In 1875, Caley published a short note on his enumeration of alkanes utilizing a tree structure \cite{cayley1875ueber}. Though Caley's enumeration ended up having a few errors, it is a very early account of treating chemical space as a structured mathematical object \cite{rains1999cayley}. Over 100 years later, the ideas of enumerating structurally similar compounds and comparing their activity became known as quantitative structure relationship studies (QSAR/SAR). QSAR/SAR is the standard method in medicinal chemistry for taking an interesting chemical compound to an optimized and potent drug lead. In 1984, Klopman developed Computer-Automated Structure Evaluation (CASE), which  "perform[s] automatically all operations related to the structure-activity analysis" \cite{klopman1984artificial}. A success in its own right, CASE utilized the graph topology of molecules to generate QSAR studies or predict activity based on fragments. This graph structure naturally leads to studying subgraphs and their relations, such as decomposing the graph into a class of similar molecules sharing a framework (scaffold), linkers connecting rings, and sidechains \cite{bemis1996properties}. Utilizing these ideas, various tool-kits and genetic algorithms have been designed to combine or grow molecular fragments into optimized drugs \cite{lameijer2005evolutionary,cernak2016medicinal}. While these ideas in organization lay the framework for certain practices of medicinal chemistry, the methods do not address the problem of enumerating compounds in an organized way to find diverse chemical scaffolds. 

Deep learning (DL) offers a new set of tools and algorithms for generating novel molecular pieces. With the introduction of generative models which can be sampled, such as variational autoencoders \cite{kingma2014semi}, or generative adversarial networks \cite{goodfellow2014generative}, de-novo molecular generation took hold as a practice in drug discovery \cite{olivecrona2017molecular}. Molecules were embedded into a continuous representation and then given a decoder, sampled from continuous space---allowing property optimization and molecular generation based on some distance metric in the latent representation. These approaches have had much success. We extend on this work by focusing on computational organization and enumeration specifically---seeking more structure than $\mathcal{N}({X}, {\Sigma})$ or $\mathbb{R}^n$.

Our contribution in this paper is twofold: (1) representing large molecular libraries using molecular building blocks (i.e., fragments, scaffold, linkers/decorations), and (2) learning to navigate latent representations of molecular hypergraphs leveraging transformer networks to \emph{operate} on molecular building blocks to generate new molecules. This project is distinct from prior molecular generation problems as we focus on enumerability and organization over property prediction. We demonstrate that our building blocks representation provides a natural mechanism to organize large chemical spaces in a statistically meaningful manner. Further, the transformer networks suggest the design of novel molecules that can `expand' on a given scaffold design, which can be used for subsequent rounds of virtual screening studies.

\section{Chemical Space as a Structured Domain}
\begin{figure}[t]
    \centering
    \includegraphics[width=0.85\columnwidth]{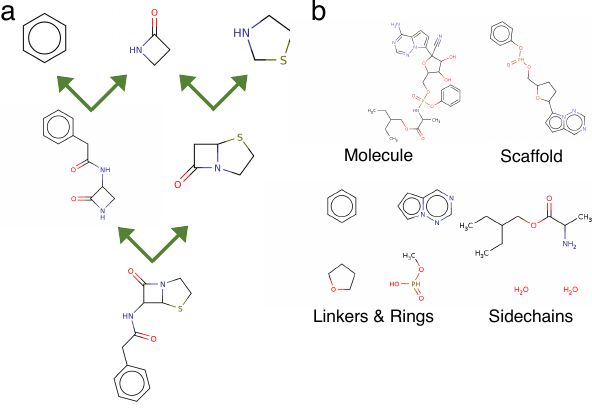}
    \caption{\textbf{Decomposition of a chemical scaffold and molecule}. (a) Starting at the bottom, a chemical scaffold with three rings is decomposed into two 2-ring scaffolds, which can be decomposed further into 1-ring scaffolds. (b) Remdesivir is decomposed from the molecule, to its scaffold, rings and linkers, and sidechains.}
    \label{fig:tree}
\end{figure}

Given the significance and applications of molecular design, several approaches have explored how to organize the chemical design space, including the use of strings (e.g., SMILES), molecular/chemical descriptor data (e.g., Modred features), and molecular graphs for both representing and generating new molecular designs. Each representation presents some opportunities and challenges in capturing the complexity of the chemical landscape and has successfully designed new molecules for drug targets and new catalysts, and other materials. However, no single representation can sufficiently capture the diversity (in chemical species) and the statistical diversity in their representations. For e.g., while SMILES strings provide a convenient means to encode chemical information, two nearly identical molecules can have significantly different SMILES representations, presenting unique challenges when embedding them into a latent manifold. Similar observations can be made for other molecular representations. 

A reemerging principle in small molecule-based property prediction models is the similar property principle \cite{johnson1990concepts}. This principle has been widely applied in the context of determining quantitative structure-activity relationships (QSAR) in medicinal chemistry: how compounds and their activities (against a drug target) can potentially improve (or degrade) based on modifying certain chemical scaffolds (or addition/deletion of R-groups)~\cite{Maggiora2011,Guha2011}. 

Consider the set of drug-like molecules $\mathcal{M}$. $\mathcal{M}$ is not directly computable as it is a concept class for \textit{molecules}. For computation, molecules require a computable representation, and this is the start of the difficulty. Representations are models of molecules which can be identified with a molecule. Graphs are a natural model of molecules, where nodes are atoms and edges are vertices \cite{kearnes2016molecular}. SMILES are another representation of molecules, which are a breadth-first search over the graph in a particular syntax. SMILES, unlike graphs, are not injective over molecules (if two SMILES strings are not equal, it does not imply the underlying molecules are not equivalent) \cite{o2012towards}. There are other representations which are less common such as point clouds, junction trees, or voxelization \cite{elton2019deep}. We define $R_X$ to be a general representation mapping from molecules to some set $X$ from $\mathcal{M}.$ 

Embeddings are distinct from representations. Embeddings are functions which take a representation $X$ to embedding space $Y$. For instance, molecular fingerprints are an algorithm which takes graphs of molecules to $\mathbb{R}^n$ by utilizing a hashing function around the nodes or regions of a graph \cite{stepivsnik2021comprehensive}. Node2vec models take graphs to $\mathbb{R}^n$. A simple variational autoencoder's encoder can take SMILES to a Gaussian unit ball $\mathcal{N}(X,\Sigma)$. The junction tree variational autoencoder takes a junction tree to a latent unit ball. In the later two examples, the idea of sampling from a normal unit ball is essential for maintaining the density of the sampling space---an important aspect of creating a generative model (see SI section 2 on sampling). Given a decoder, these embedding spaces can be sampled to produce potentially new molecules or molecules through a constrained optimization problem. The two embedding spaces so far have convenient distance metrics, denoted $\delta_Y$. 

A number of papers have focused on generative models for the design of new molecules~\cite{duvenaud2015convolutional,kearnes2016molecular,gilmer2017neural,gomez2018automatic,kusner2017grammar}. These approaches either use a string representation (e.g., SMILES representation mapped onto a molecular graph) or an explicit molecular graph representation (e.g., \cite{jin2018junction}) to \emph{encode} the molecular data into a continuous representation from which new examples can be drawn. 

While these methods are very successful at certain property predictions and general optimization, they do not solve the enumerability problem. Both $\mathbb{R}^n$ and $\mathcal{N}(X,\Sigma)$ are continuous and not countable. In particular, every molecule has an open ball around it in embedding space of equivalent points which is a problem for enumerating discrete sets of molecules. In other words, if $\varphi^{-1}$ is a decoder from an embedding $\mathbb{R}^n\rightarrow X$, and $\equiv$ is an equivalence relation on the representation $X$, there exists $y_1,y_2\in\mathbb{R}^n$ and $\epsilon>0$ such that $0<\delta_{\mathbb{R}^n}(y_1,y_2)<\epsilon$ so
\begin{equation*}
\varphi^{-1}(y_1)\not\equiv \varphi^{-1}(y_2).
\end{equation*}
In order to structure the embedding space to be conducive for enumeration, we must find an embedding space that is countable and discrete, just as Caley sought out by means of a tree. 

Molecular scaffolds are well defined through algorithms, decompose well into networks, and offer a general description of global properties (such as orientation in a protein binding region) \cite{bemis1996properties, 10.1093/bioinformatics/btaa219}. Molecular scaffolds represent the core of a molecule, typically defined around the number of rings in the structure. Non-ring sturctures in molecules include linkers and sidechains which get collapsed in this representation to a single scaffold representative. In figure \ref{fig:tree}, we show a molecular scaffold decomposing into smaller scaffolds. In this way, we can take a graph or SMILES representation of a molecule and map it to this discrete embedding structure. The mapping into the scaffold structure is unique. As other authors rely on decoders to decode the embedding space, we will rely on decoders to sample the scaffold for the variety of molecules a part of it.

\section{Methods}

\begin{figure}[t]
    \centering
    \includegraphics[width=0.47\textwidth]{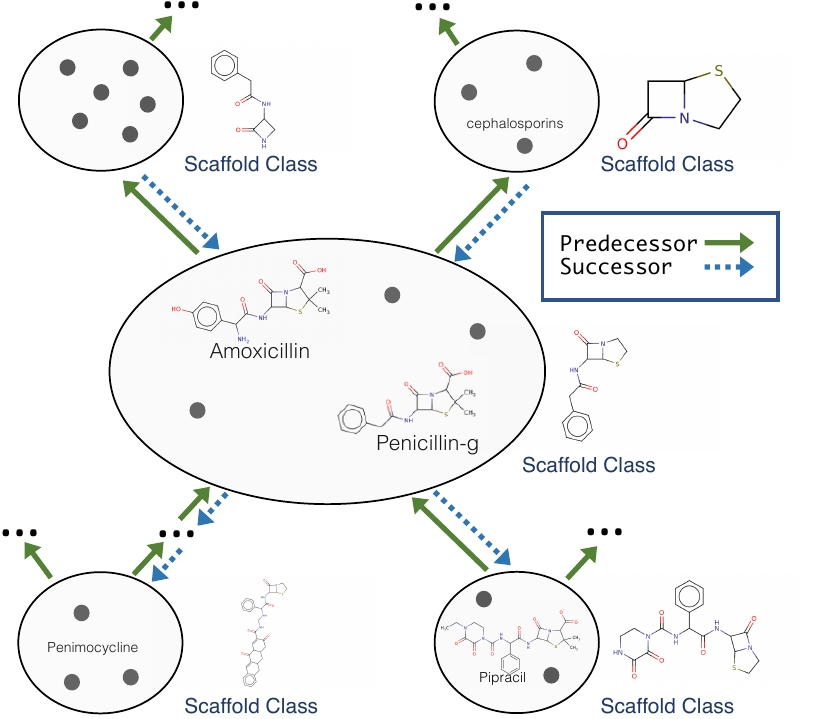}
    \caption{\textbf{Scaffold as classes over drug-like chemical space.} Every molecule (represented by dots or depiction inside circles) is inside a single scaffold class. Scaffold classes are related through common substructures, forming a hierarchy of classes. Penimocycline, for example, belongs to a scaffolding class from far Penicillin-g's or Amoxicillin's class, while Pipracil is a direct successor of the Penicillin-g class. The \Predecessor{} function is defined via an algorithm, and the \Successor{} function requires a generative model when working without data (i.e., given a single scaffold you cannot compute its successor unless you understand chemistry, thus have parameters $\Phi$, but you can compute all of it is predecessors recursively without knowing how to generate new compounds).}\label{fig:workflow}
\end{figure}

The conceptual machinery for treating chemical space as a hypergraph structured through scaffolds is developed. There is an elegant statement of the principle of fragment-based drug design through the operations among scaffolds. Further, the framework developed provides intuitive concepts for understanding the diversity and size of chemical space explored or discussed by a model or computational research program. As a computational learning problem, we use transformer as seq2seq models to implement large graph navigation in practice.

\subsection{Scaffold Embeddings}
Utilizing the concept of scaffolds developed in section 2, we assume the operation \Scaffold{} as a given oracle such that \Scaffold{} is injective and defined for every molecule. We define $\mathcal{S}$ as the set of all scaffolds. 

A hypergraph is a generalized graph where edges group more than two vertices. A hypergraph is $n$-regular when every vertex is contained in exactly $n$ edges. Scaffolds as hypergraph edges over molecules form a $1$-regular graph, as every molecule belongs to exactly one scaffold class, thus every vertex has degree 1 in the hypergraph. We denote the hypergraph as $\mathcal{H}=(\Mspace, \Sspace)$. 

\begin{figure}[t]
    \centering
    \includegraphics[width=0.7\columnwidth]{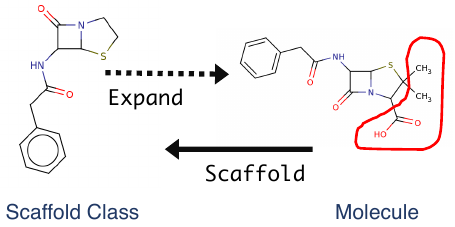}
    \caption{\textbf{Scaffold and molecule relation}. Scaffolds are the core or framework of a molecule, and they represent a class of molecules. Scaffolds, or scaffold classes as we often refer, group molecules together. A class can be extended by adding decorations to the scaffold, such as linkers and sidechains. Through the scaffold function, we obtain the scaffold of a molecule.}
    \label{fig:expand}
\end{figure}

\textbf{Operations on scaffolds}.
We denote computational operations in \texttt{Monospace} font, and add a subscript $\Phi$ to represent parameters which may be required for the operations (i.e. \Expand{}). 
\begin{enumerate}
    \item \Expand{} and \Scaffold{}: Molecules and scaffolds represent two distinct types which can be converted back and forth (figure \ref{fig:expand}). Scaffold classes can be \textit{expanded}, where we envision zooming in, via the \Expand{} model (i.e. \Expand{}$:\Sspace\rightarrow\Mspace$). Similarly, molecules can be taken to their scaffold via the program \Scaffold{} (\Scaffold{}$:\Mspace\rightarrow\Sspace$). We utilize RDKit to compute \Scaffold{} via the MurckoScaffold module \cite{landrum2016rdkit}. We note a model can be trained for this task; however, given the efficiency of the algorithm it did not seem fruitful at this time. 
    \item \Successor{} and \Predecessor{}: the successors of a scaffold $S_1$ are the set of all scaffolds $S$ which contains $S_1$ as a substructure (figure \ref{fig:workflow}). The predecessors of a scaffold $S_1$ are all scaffolds $S$ which $S_1$ is a superstructure. In general, there is no algorithm for successor given only a scaffold, as it requires sampling chemical space. However, predecessor has an efficient algorithm with a structure that can always be fragmented into smaller scaffolds without sampling other data. These operations are the atomic building blocks of navigating between scaffold classes (and induces a strict partial ordering ($\mathcal{S}$, $\prec$)). These operations are from $\Sspace$ to $\Sspace$. We also consider the standard graph structure induced by the relation \Successor{} and \Predecessor{}, and denote it $\mathcal{S}_\mathcal{G}=(\mathcal{S}, $\Successor$)$ where \Successor{} can be used to determine the edge relation. This graph can be directed or undirected, but for our case we consider the undirected graph mostly.  
    \item \Union{} and \Intersection{}: two scaffolds $S_1$ and $S_2$ can be combined to form a union. More formally, the union of $S_1$ and $S_2$ is the set of scaffolds that contain $S$ where $S$ has $S_1$, and $S_2$ has immediate predecessors. Similarly, the intersection of $S_1$ and $S_2$ is simply the maximum common substructure (MCS) of $S_1$ and $S_2$, for which an efficient algorithm exists for small drug-like molecules.  \cite{cao2008maximum, cone1977computer}. In general, MCS is NP-complete, but there are heuristics for drug-like molecules that provide a rather efficient algorithm \cite{garey1979computers}. These operations are from $\Sspace$ to $\Sspace$.
\end{enumerate}

These basic operations can be combined into more complex operations such as \begin{align}
    &\text{\texttt{UpperCone}}_{\Phi}(S)=\{A : S\prec A\} \\
\quad\text{or}\quad &\text{\texttt{LowerCone}}(S)=\{B : B\prec S\}
\end{align}
Upper cones of scaffold classes are actually a common object of interest for drug discovery. For instance, Penimocycline is in the upper cone of Penicillin-g's scaffold class (see figure \ref{fig:workflow}). Successful exploration of upper cones is the theoretical cornerstone of fragment based drug design \cite{schiebel2016high, murray2010structural}. Recently, fragment X-ray crystalgraphic screens have been performed on important drug targets such as SARS-CoV-2 proteases in search of an inhibitor \cite{douangamath2020crystallographic}. Given a set of fragment hits for a protein target in a binding region,$\{m_i\}_{i\in H}$ , take the scaffold classes of those hit, $\{S_i^h\}_{i\in H}$. The principle of fragment based drug design can be expressed as there exists some index set $I^*$ such that $I^*\subseteq H$ and
\begin{equation} 
\hat{H}=\bigcap_{i\in I^*} \text{\texttt{UpperCone}}_{\Phi}(S_i^h)
\end{equation}
where $\hat{H}$ is a set of scaffold classes, $\hat{H}$ is not empty, and some molecule in a scaffold in $\hat{H}$ is a likely candidate. In other words, a set of fragments can be grown to sets of larger drug-like molecules, and some intersection of those possible larger molecules will be a hit that is likely a drug lead for this protein target. In an embedding space such as $\mathbb{R}^n$, the same principal does not apply, and is dependent on the embedding context (for instance, based on a particular property \cite{iovanac2019improved}). Furthermore, there no guarantees about molecules in an interval between two molecules, whereas the intersection of upper cones, for example, does have such guarantees (if it is not empty). 

Given there is no algorithm for producing successor scaffolds without relying on sampling chemical space, we treat the problem as a learning problem. We note that we cannot rely on fragments as a vocabulary given this construction as other methods have (for instance, \cite{jin2018junction} utilized a finite vocabulary containing one member rings, linkers, and sidechains from the dataset). When using such a vocabulary, there are chains of scaffolds that cannot be represented as the \Successor{} function can only sample scaffold classes $S$ for which every one ring member in the \texttt{LowerCone}($S$) is in the finite vocabulary.

\subsection{Modeling Hypergraphs with Transformers}

While the method outlined has no constraints on compounds' synthetic accessibility, it is a necessary and essential aspect of chemical space exploration for drug discovery. To focus on synthetic accessibility while paying attention to maximizing library size, we utilize a dataset from Synthetically Accessible Virtual Inventory (SAVI) \cite{patel2020savi}. SAVI contains over 1.7 billion reaction products (along with rich reaction and metadata). We utilize only the SMILES of the products. 

We build two datasets from SAVI. The first utilizes RDKit to determine the scaffold for each of the compounds listed \cite{landrum2016rdkit}. We utilized a 200M sample from the entire dataset and extended the data by a factor of 5 by randomizing the SMILES both for the target (scaffold) and source (molecule) \cite{arus2019randomized}. A set of 20M molecules with a unique scaffold class are held out as validation data. A second dataset is created by taking a subsample of the prior dataset, 20M, and utilizing the ScaffoldGraph package to decompose each scaffold into a network of scaffolds \cite{10.1093/bioinformatics/btaa219}. We sample edges (representing the successor of two scaffold nodes), resulting in a dataset of five million successor pairs. This dataset is extended to 50M utilizing random smiles sampling. Predecessor data is flipping the columns (sources become targets, and targets become sources) for the successor datasets.

On the one hand, \Successor{} and \Expand{} are generative models---given a scaffold, those operators are required to sample the space of successor scaffolds or molecules that have that scaffold. On the other hand, they are seq2seq task, taking one sequence to a different sequence. This combination of wanting a dense sampling strategy combined with seq2seq modeling differs from applications we have found in the literature. Common approaches to generative models have been utilizing VAEs or GANs to train some encoder-decoder model on sample reconstruction error with some regularization \cite{elton2019deep, gupta2018generative, grisoni2020bidirectional}. Seq2seq approaches in this space have focused on solving problems with a relatively small optimal solution set such as reaction modeling \cite{schwaller2019molecular}. With the recent success of transformer models performing well on large datasets and seq2seq problems, we decided to follow the modeling as a seq2seq problem as Schwaller et al. have. 

 We utilize a transformer seq2seq model from the ONMT project \cite{klein-etal-2017-opennmt}. Other works have utilized RNNs, but we utilize a transformer for both the encoder and decoder of the model \cite{vaswani2017attention}. Given the goal of not simple generation but rather generalizing a very large hypergraph for which a pure algorithmic solution is intractable, transformer models are a good fit compared to simpler RNN models. Code is compiled into a GitHub repository with scripts for data gathering, data preparation, model training, and sampling. The interface is geared towards developing front-end functions for quick medicinal chemistry questions regarding sampling molecular space.

\section{Experiments}

\subsection{Computability of Scaffold Classes}
\label{covering}

\begin{figure}[t]
    \centering
    \includegraphics[width=0.95\columnwidth]{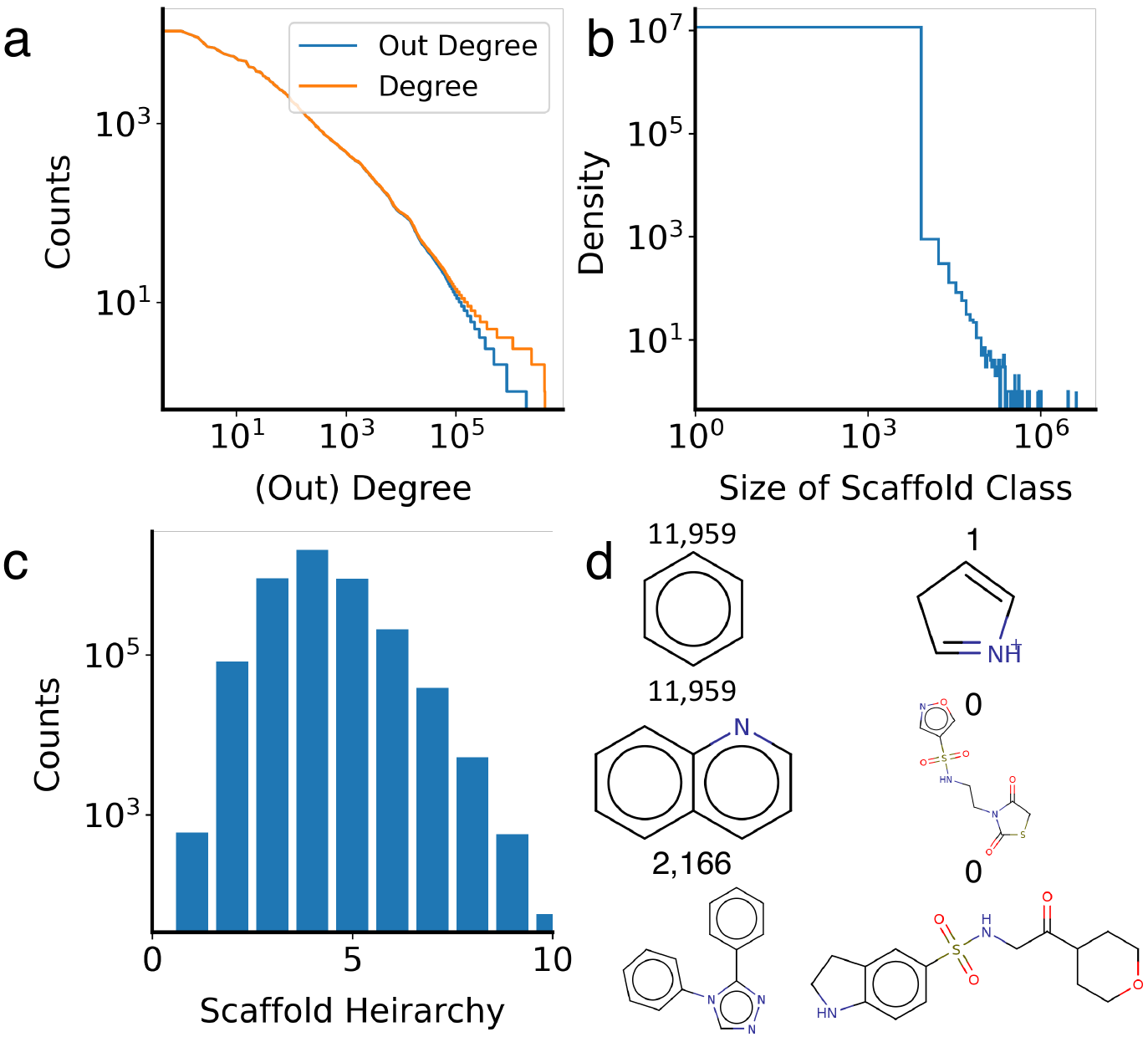}
    \caption{\textbf{Structure of scaffold classes} We constructed the scaffold classes (4M) for a random sample from SAVI (20M) molecules for (a)-(d). (a) We consider a random sample of 20M molecules from SAVI, and construct the scaffold classes and graph associated with the classes. Out degree indicates just \texttt{Successor} relations. (b) We show the distribution of the cardinality of (a)'s scaffold classes, which follows a power law for part of the distribution, and a uniform distribution for the other. (c) Scaffold classes are ordered into a hierarchy based on the number of rings its framework has. (d) The left column shows the scaffolds with the largest out degrees for hierarchies 1 to 3, and the right column shows random scaffolds of the least degree. }
    \label{fig:density}
\end{figure}

We assess the structure of \textit{scaffolding} chemical space, focusing on understanding the size of scaffold classes, how many scaffold groups there are in drug-like chemical space, and how they connect. 

We impose a structure on $\mathcal{M}$ by creating scaffold classes $\mathcal{S}=\{{S}\}_{i\in I_s}$ such that every molecule $m$ belongs to one and only one scaffold class, and all classes in $\{{S}\}_{i\in I_s}$ are disjoint. We also assign a hierarchy to scaffolds based on the number of rings. $\mathcal{H}_n$ is the set of all scaffold classes with ring size $n$.  

$\mathcal{H}_0$ is the smallest hierarchy, which consists of only one scaffold class ${S}_0$, the set of all molecules with no rings (ring-less fragments, linkers, and side-chains). $\mathcal{H}_1$ is the set of all scaffold classes with one ring. The order of $\mathcal{H}_2$ is proportion to $\vert\mathcal{H}_1\vert$ choose $2$ plus the combination of linkages and sidechain modifications from $\mathcal{H}_0$. We see growth similar to the partition function in theory. However, in practice, the distribution of molecules in real-world datasets typically follows a normal distribution with the mean around three rings (see figure \ref{fig:density}). 

Given this added structure of scaffolds, {do scaffolds reduce the search space over molecules by many magnitude orders?} If this is the case, we can search through a computable number of scaffolds, and once a few interesting classes are found, we can enumerate the molecules in that set. This strategy does not face the curse of $10^{68}$ drug-like molecules the current unstructured domain $\mathcal{M}$ faces. Given a 200M sample from SAVI, we found only 11.4M (5.7\%) scaffold classes were needed to cover the entire dataset, and, in practice, there exists a large subset of molecules (165M) with only 685,000 (0.41\%) scaffold classes. This reduction via scaffolds implies for a large subset of molecules, there is a reasonable 5 order of magnitude gain in search over scaffolds than pure molecules (from a database or chemical library perspective). 
\subsection{Hyerpgraph Navigation}

We train three operations (\Expand{}, \Successor{}, \Predecessor{}) utilizing three separate models. While there is an algorithm for \Predecessor{}, we can compare it directly to the algorithm performance. Each model was trained for approximately two days on eight GPUs (NVIDIA Tesla V100). Each model was trained for 500,000 steps with a batch size of 8192. Further details of the training procedure can be found in (SI) and on GitHub.\footnote{Upcoming} 

\begin{table}[t]
\centering
\label{tab:acc}
\resizebox{\columnwidth}{!}{%
\begin{tabular}{@{}lccc@{}}
\toprule
Model      & SMILES Validity & Type Accuracy & Correctness Accuracy \\ \midrule
\Successor{} &   98.9\%              & 98.9\%             &  97.9\%                  \\
\Predecessor{} &        99.8\%         &   99.8\%            &      94.0\%                \\
\Expand{} &  98.6\%               &    -           &   96.9\%                    \\ \bottomrule
\end{tabular}
}
\caption{Performance metrics from graph navigation models. Evaluations were performed with a holdout set from SAVI dataset. SMILES validity is the percent of samples that pass an RDKit parser. Type accuracy determines how many samples have the correct type (\texttt{Successor}, \texttt{Predecessor}, and \texttt{Union} models output type scaffold. In contrast, \texttt{Expansion} model outputs molecules (which can include a scaffold representative, and this metric is left out and computed as a part of correctness). Correctness accuracy is the percent of samples which are valid, typed correctly, and are equivalent to the algorithmic solution.} 
\end{table}

\begin{figure}[t]
    \centering
    \includegraphics[width=0.49\columnwidth]{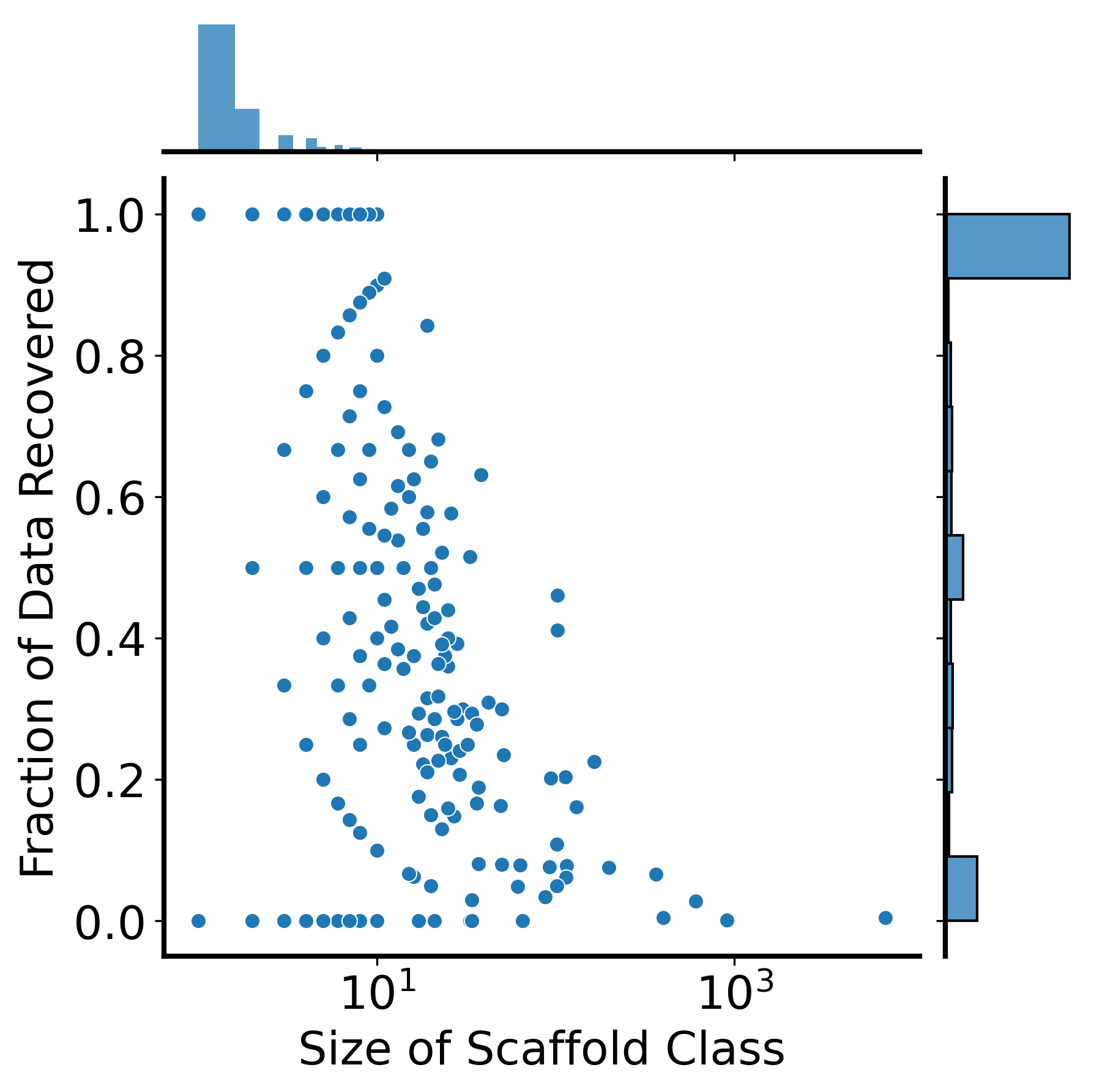}
    \includegraphics[width=0.49\columnwidth]{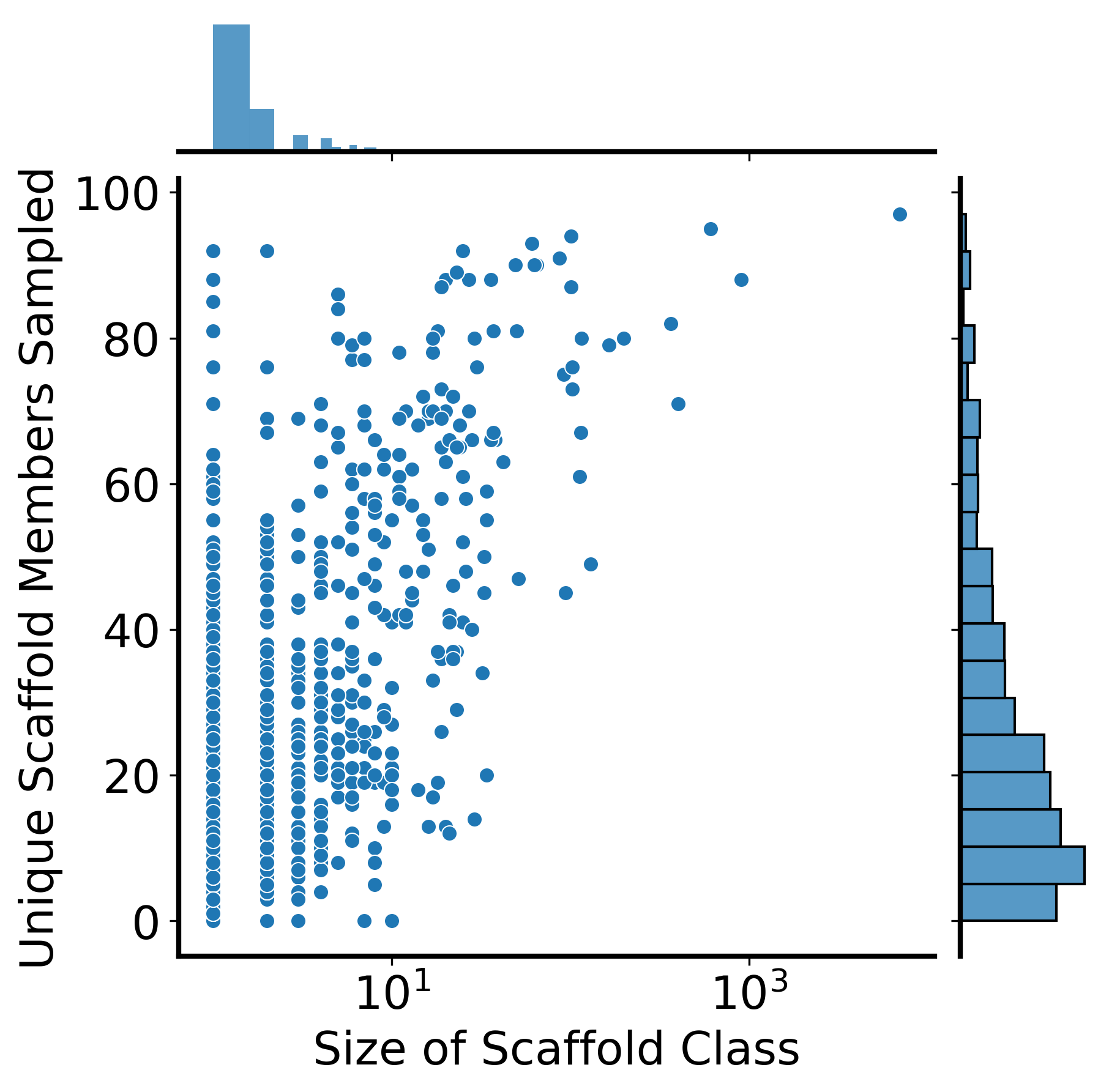}
    \caption{\textbf{\Expand{} model reconstruction and sampling depth.} 1000 samples scaffold classes are drawn from the validation data, and \Expand{} is sampled 100 times. Samples that are not valid smiles or passed verification are removed. (\textit{left}) Samples for each scaffold are intersected with the known molecules in that scaffold class from the validation data, and the fraction found is plotted. Smaller scaffolds are often recovered while larger ones are not. (\textit{right}) Even though the \Expand{} model captures most of the dataset for smaller scaffolds, the model generates more valid molecules based on the natural distribution of the scaffold class sizes in the data.}
    \label{fig:recovery}
\end{figure}

To sample \Expand{} and \Successor{} we utilize beam search with a temperature of 1.5, beam size of 5, and randomizing the SMILES input. Samples are then validated utilizing RDKit. In table \ref{tab:acc}, we outline each model's accuracy. A uniform sample of scaffolds from the validation data was taken ($n=1000$), and 100 samples were drawn for each scaffold class (figure \ref{fig:recovery}). 

Given the density of some scaffold classes in the data compared to others (figure \ref{fig:density}), more advanced sampling methods required for \Expand{} on these classes. For scaffold classes with over $10^6$ members in the data (mostly 1-ring and 2-ring common scaffolds), resampling validation data from the model is difficult (table \ref{tab:dense}). Given the uniqueness of sampling based on a category like scaffolds, rather than pure sampling points in a distribution or $\mathbb{R}^n$, comparisons to generative models' reconstruction accuracy are not reasonable. 

\begin{table}[t]
\centering
\label{tab:dense}
\resizebox{\columnwidth}{!}{%
\begin{tabular}{@{}lccc@{}}
\toprule
Scaffold                            & Class Size (Data) & Unique Sampled & Overlap (Recall) \\ \midrule
c1ccc(COc2ccccc2)cc1                & 373,939            & 168,261         & 4,146 (1.1\%)     \\
O=S(=O)(c1ccccc1)N1CCCCCC1          & 88,608             & 145,904         & 20,097 (22.7\%)   \\
O=S(=O)(NCCc1ccccc1)c1ccccc1        & 911,360            & 176,539         & 23,715 (2.6\%)    \\
c1ccncc1                            & 818,230            & 183,838         & 23,999 (3.0\%)    \\
O=S(=O)(NS(=O)(=O)c1cccnc1)c1ccccc1 & 203,891            & 173,599         & 20,331 (10.0\%)   \\ \bottomrule
\end{tabular}
}
\caption{\textbf{Sampling dense classes with \Expand{}}. Five dense scaffold classes were taken from the validation data and sampled. We sampled 100,000 times for each scaffold, utilizing a temperature of 1.5 and a beam search of length five and capturing the top two best beams from the search. While we do not capture a large set of the data, we believe these classes' sheer size presents a combinatorics problem. The unique samples are all correct and valid.}
\end{table}

\begin{figure}
    \centering
    \includegraphics[width=0.75\columnwidth]{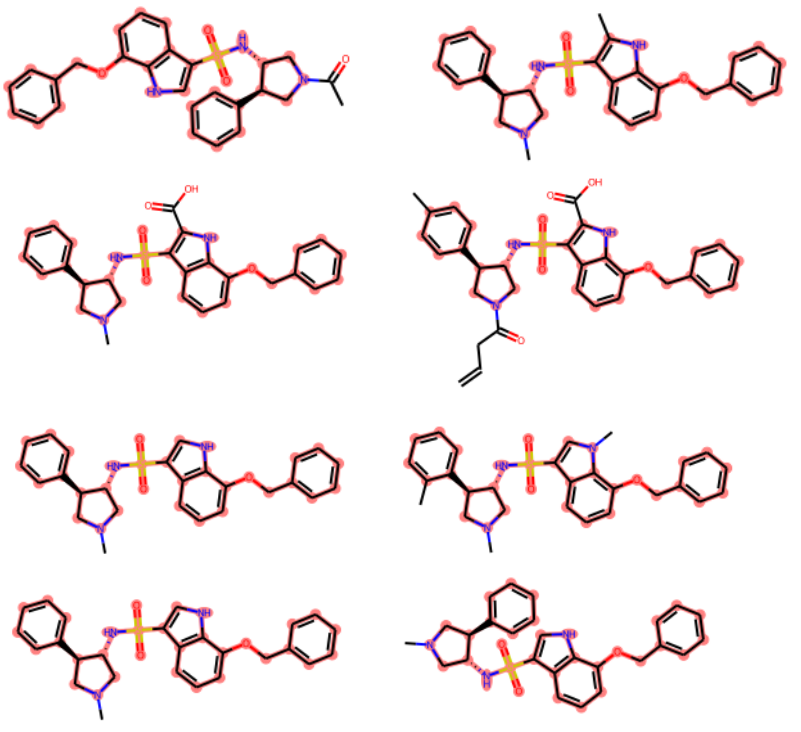}
    \caption{\textbf{Expansion of a scaffold.} The expansion of a scaffold class, highlighted in red, is expanded by sampling \Expand{}. Various side chains are added, but no sample is outside of the class.}
    \label{fig:qsar}
\end{figure}

Figure \ref{fig:qsar} is an example of a series of compounds which belong to a single scaffold class, but are sampled with different sidechains. The variety of sidechains while maintaining the single scaffold core is the basis of a QSAR series. 

\section{Conclusion}
This paper outlined a set of ordered equivalence classes via molecular scaffolds over the drug-like chemical space (forming a 1-regular hypergraph). We utilize seq2seq models to move between scaffolds, or classes of compounds, and between the scaffold hierarchy and the underlying molecules themselves. These operations ultimately form a set of algebraic tools for manipulating and navigating the chemical space. This algebra is expressive---enough to represent algorithms in drug design, such as the principle of fragment-based drug design or similar property principle of molecular scaffolds. This construction over $\mathcal{M}$ offers a unique take on the enumerability of the chemical space by collapsing the space into scaffold classes, which can zoomed-in or zoomed-out of. We aim to understand better the distribution of synthetically accessible drug space and its relation to scaffolds as we hope scaffold classes reduce the space's overall size. Future work will unify the algebra into a single model for navigating the space and introduce more concept classes for finer and coarser granularity. We believe that to accelerate exploring the estimated $10^{60}$ drug-like molecules, a navigation strategy besides standard databases and compound enumeration is needed.

\normalsize
\bibliography{main}










\end{document}